\documentclass{article} 
\usepackage{iclr2026_conference,times}

\usepackage{amsmath,amsfonts,bm}

\def\eqref#1{equation~\ref{#1}}

\def\1{\bm{1}}

\DeclareMathAlphabet{\mathsfit}{\encodingdefault}{\sfdefault}{m}{sl}
\SetMathAlphabet{\mathsfit}{bold}{\encodingdefault}{\sfdefault}{bx}{n}

\usepackage{hyperref}
\usepackage{url}

\usepackage{amssymb}
\usepackage{booktabs}
\usepackage{tikz}
\usepackage{xspace}
\usepackage{arydshln}
\usepackage{enumitem}
\usepackage{nccmath, amsmath}

\DeclareRobustCommand*\circled[1]{\tikz[baseline=(char.base)]{ \node[shape=circle,draw,color=white,fill=black,inner sep=0.5pt] (char){{\footnotesize{#1}}};}}

\DeclareRobustCommand*\circledRed[1]{\tikz[baseline=(char.base)]{ \node[shape=circle,draw,color=white,fill=red,inner sep=0.5pt, font=\bfseries] (char){{\footnotesize{#1}}};}}

\def\ie{{i.e.},~}
\def\eg{{e.g.},~}

\newcommand{\shortsectionBf}[1]{\noindent {\bf #1}}

\newcommand{\hostagent}{{\textsc{\small{HA}}}\xspace}
\newcommand{\llm}{{\textsc{\small{LLM}}}\xspace}
\newcommand{\hostagentcore}{{\textsc{\small{HAC}}}\xspace}
\newcommand{\sessionmanager}{{\textsc{\small{SM}}}\xspace}
\newcommand{\dialoguemanager}{{\textsc{\small{DM}}}\xspace}
\newcommand{\communicationlayer}{{\textsc{\small{CL}}}\xspace}

\newcommand{\externalentity}{{\textsc{\small{EE}}}\xspace}
\newcommand{\validationmodule}{{\textsc{\small{VM}}}\xspace}

\setlist[itemize]{leftmargin=*, itemsep=0.2pt, topsep=0.2pt} 
\setlist[enumerate]{leftmargin=*, itemsep=0.2pt , topsep=0.2pt} 
\usepackage{caption} 

\title{Formalizing the Safety, Security, and Functional Properties of Agentic AI Systems}

\author{Edoardo Allegrini \\
Sapienza University of Rome \\
\texttt{allegrini@di.uniroma1.it} \\
\And
Ananth Shreekumar \\
Purdue University \\
\texttt{ashreeku@purdue.edu} 
\And Z. Berkay Celik \\
Purdue University \\
\texttt{zcelik@purdue.edu}
}

\iclrfinalcopy 
\begin{document}

\maketitle

\begin{abstract}
Agentic AI systems, which leverage multiple autonomous agents and large language models (LLMs), are increasingly used to address complex, multi-step tasks. The safety, security, and functionality of these systems are critical, especially in high-stakes applications. However, the current ecosystem of inter-agent communication is fragmented, with protocols such as the Model Context Protocol (MCP) for tool access and the Agent-to-Agent (A2A) protocol for coordination being analyzed in isolation. This fragmentation creates a semantic gap that prevents the rigorous analysis of system properties and introduces risks such as architectural misalignment and exploitable coordination issues. To address these challenges, we introduce a modeling framework for agentic AI systems composed of two central models: (1) the \emph{host agent} model formalizes the top-level entity that interacts with the user, decomposes tasks, and orchestrates their execution by leveraging external agents and tools; (2) the \emph{task lifecycle} model details the states and transitions of individual sub-tasks from creation to completion, providing a fine-grained view of task management and error handling. Together, these models provide a unified semantic framework for reasoning about the behavior of multi-AI agent systems. Grounded in this framework, we define 16 properties for the host agent and 14 for the task lifecycle, categorized into liveness, safety, completeness, and fairness. Expressed in temporal logic, these properties enable formal verification of system behavior, detection of coordination edge cases, and prevention of deadlocks and security vulnerabilities. Through this effort, we introduce the first rigorously grounded, domain-agnostic framework for the analysis, design, and deployment of correct, reliable, and robust agentic AI systems.
\end{abstract}

\section{Introduction}
Agentic AI systems are becoming central to the creation of advanced AI. These systems address complex, multi-step tasks that surpass single-agent capabilities by utilizing multiple autonomous agents. Agents in these systems employ Large Language Models (LLMs) as their primary reasoning engine, which enables them to perform task planning, delegation, and complex decision-making~\citep{simulacra, llm_mas_adaptive, cooperative_llm_mas}. These heterogeneous agents (\eg agents based on language, vision, robotics, and symbolic planning) dynamically communicate, delegate, and collaborate to complete tasks. 

To illustrate, we consider an automated financial planner. A \emph{host agent} accepts a natural language request from a user to ``create a budget and invest $\$1,000$''. It decomposes this into subtasks: a \emph{data agent} gathers market data, a \emph{planning agent} calculates risk, and a \emph{transaction agent} executes the final investment. These specialized agents interact and exchange data to accomplish the multi-step task. The operation of such systems relies on standardized communication protocols, such as  Model Context Protocol (MCP)~\citep{anthropicMCP}, which addresses vertical tool access for a single agent, and  Agent-to-Agent (A2A)~\citep{googleA2A} protocol, which addresses horizontal agent coordination via inter-agent delegation. 

Turning back to the financial planner example, the agentic system must leverage both protocols. The \emph{host agent} uses A2A to delegate the subtask of gathering market data to the \emph{data agent} (horizontal agent coordination). The \emph{transaction agent} relies on MCP to invoke an external banking API tool to execute the final investment (vertical tool access). The host agent's ability to manage and sequence both A2A and MCP interactions across various agents enables the multi-step task to succeed.

Formal or empirical assessment of the \emph{safety}, \emph{security}, and \emph{functionality} of multi-AI agent systems is critical as these systems transition to open-ended, high-stakes applications~\citep{dewitt2025openchallengesmultiagentsecurity, llmmultiagentsystemschallenges}. Here, safety involves preventing the agent's actions from causing unintended or harmful physical or real-world consequences, such as an incorrect trade that causes a substantial loss in the financial planner example. Functionality ensures that the system accurately and reliably completes the specified task, such as verifying that the financial planner calculates risk correctly and executes the investment as requested. Safety and functionality issues may arise naturally, \eg due to design flaws or code errors, whereas security considers adversarial manipulation and unauthorized access that can violate these requirements. For instance, a malicious external agent could use a prompt injection attack~\citep{liu2024formalizing} to force the \emph{transaction agent} into transferring funds to an unauthorized account, a clear violation of both security and safety policies.

Recent work directly or indirectly studies these issues by examining agents in specific settings, such as adversarial agent behavior~\citep{maliciousagents, redteamingllmmas, massecurityTAX} and protocol misuse in task delegation~\citep{motwani2024secret}. Others focus on the compromise or malfunction of a single agent that propagates errors, escalates failures, or exposes vulnerabilities across an entire system~\citep{cascade_llm_compromise, llm_to_llm_PI, physical_safety_MAS_noAI, security_in_mas2015}. Further studies examine the risk of using combinations of safe models for misuse~\citep{combination_misuse_adversaries}, the security challenges inherent in tool usage~\citep{toolsword}, and ensuring secure delegation and preventing protocol misuse in adversarial settings~\citep{buildingsecurea2a, li2025gluecodeprotocolscriticalanalysis}.

The current ecosystem of inter-agent communication and coordination is fragmented and lacks a cohesive formal foundation. Existing protocols (\eg MCP and A2A) are analyzed in isolation, which creates a semantic gap that impedes rigorous reasoning about safety, security, and functional properties when both are used simultaneously in complex workflows. This absence of a unified framework, a need that is also recognized by multi-stakeholder initiatives~\citep{neelou2025a2as}, leaves deployments vulnerable to emergent coordination failures, including deadlocks and privilege escalation, that cannot be adequately verified or mitigated. Achieving verifiable correctness in these systems requires a rigorous, unified modeling framework that can capture the execution behavior of diverse agents across heterogeneous protocols.

To address these challenges, we introduce a two-model framework for agentic AI systems:

\begin{enumerate}
\item The \textbf{Host Agent Model} formalizes the top-level entity that interacts with the user. It is responsible for accepting user tasks, decomposing them into structured subtasks, and orchestrating execution via external entities (agents via A2A, tools via MCP). It acts as both a controller and a monitor, ensuring safe delegation and state consistency throughout the task lifecycle.

\item The \textbf{Task Lifecycle Model} formalizes the dynamic structure of task management, detailing states and transitions a sub-task undergoes from its origin to its completion or failure. It provides the fine-grained execution logic necessary for dependency management and resilient error handling.
\end{enumerate}

These models are the foundational structure that enables formal verification. They capture discrete states and transitions of the entire coordination process and provide the necessary semantic framework (state space) to express the formal safety, security, and functional requirements. This unified perspective is required to rigorously reason about execution behavior, uncover coordination edge cases, and verify properties critical to real-world deployments. 

We categorize the formal properties defined within this framework as liveness, safety, completeness, and fairness, which are essential for verifiable system assurance. For instance, a \emph{safety} property guarantees that the transaction agent will not execute an investment before the planning agent has successfully calculated the risk, preventing inconsistent state or premature execution. Conversely, a \emph{liveness} property ensures that once a user submits a request, a final response is eventually returned. This prevents a task from entering a permanent deadlock or starvation state.

\shortsectionBf{Contributions.} In summary, we make the following contributions:

\begin{itemize}
\item We introduce a host agent model, providing a rigorous and domain-agnostic foundation for the systematic analysis of multi-agent systems. We formalize the top-level agent that interacts with users, decomposes tasks, and orchestrates their execution by leveraging external agents and tools.

\item Building upon the host agent model, we introduce the first definition of the task lifecycle model. This model unifies tool-use protocols and inter-agent communication protocols, establishing a common semantic framework for task initiation, delegation, execution, and completion. 

\item We define critical formal properties for multi-agent AI systems, derived from the host agent and task lifecycle models. These properties ensure the necessary requirements for safety, security, and functionality are satisfied regardless of the deployment domain.

\end{itemize}

\section{Background}

\shortsectionBf{Agentic AI systems.} AI agents are autonomous entities that perceive, reason, and act to achieve goals~\citep{russell2016artificial}. While early systems were symbolic~\citep{nilsson1998artificial}, modern agents use \llm{s} to reason, plan, and execute tool-use tasks~\citep{wang2024survey_llm_agents, durante2024agent_ai, llm-agents-survey}. However, complex multi-faceted tasks often exceed single-agent capabilities. Agentic AI systems address this by forming collaborative networks~\citep{stone2000multiagent, dorri2018multi}, where specialized agents interact to solve problems beyond individual scope through dynamic task decomposition and allocation~\citep{tampuu2017multiagent, li2024camel, simulacra}.

\shortsectionBf{Agent Communication and Coordination.} Agentic AI systems require standardized frameworks for individual agent capabilities and multi-agent coordination. These frameworks comprise two complementary aspects: protocols for agents to interact with external resources and tools, and protocols for inter-agent communication. For individual agents, the challenge is ensuring secure, reliable access to external computational resources and tools. Agents must discover capabilities, authenticate services, validate parameters, and handle errors robustly. For multi-agent coordination, protocols define how agents discover peer capabilities, negotiate task assignments, exchange information, and maintain consistency across distributed execution contexts.

The MCP~\citep{anthropicMCP} and A2A~\citep{googleA2A} protocols illustrate the integration challenges that arise from combining individual agent capabilities with multi-agent coordination. MCP standardizes agent-tool integration~\citep{anthropicMCP}. Operating as a client-server protocol, it enables agents to discover, authenticate and execute commands across external data sources and toolsets. This hub-and-spoke model ensures agents can retrieve context and trigger actions without requiring unique code for every provider.

While MCP facilitates tool usage, A2A enables horizontal coordination between autonomous agents~\citep{googleA2A}. Built on HTTP and JSON-RPC, A2A allows agents to discover peer capabilities and negotiate task assignments across distributed environments. Central to A2A is the \emph{Agent Card}, metadata describing an agent's capabilities and endpoints. This allows agents to establish trust and delegate tasks dynamically without shared memory.

\section{Challenges in Compositional Reasoning of Agentic AI Systems}
Integrating MCP and A2A within Agentic AI systems creates a composition of multiple components. These operate on different abstractions and assume distinct trust models~\citep{dewitt2025openchallengesmultiagentsecurity}. For instance, MCP treats external tools as trusted resources with well-defined interfaces, whereas A2A handles potentially untrusted peer agents with varying capabilities and reliability guarantees. 

\shortsectionBf{Architectural Misalignment and Semantic Gap.}
The core difficulty arises from the lack of a unified semantic layer between MCP's tool-centric model and A2A's agent-centric model. This misalignment makes it difficult to reason about correctness properties, maintain state consistency, and verify security guarantees when both protocols are used simultaneously within the same system~\citep{massecurityTAX}. The complexity is further amplified by the dynamic nature of agents, which may exhibit behaviors that violate protocol assumptions~\citep{cooperative_mas_security}. This creates a lack of formal guarantees in two areas:

\begin{itemize}
\item \textbf{Task Handoff Failures}: Transferring a task from an A2A-delegating agent to an MCP-invoking agent is prone to failure. This creates unreliable delegation and coordination, which leads to inconsistent state management and inadequate validation of delegation chains. For example, the host agent may delegate a task via A2A to a specialist agent, but the specialist agent may fail to correctly translate or format the required parameters before invoking a crucial external tool using the MCP, thus halting the execution chain.

\item \textbf{Inconsistent State Management}: Without a unified view, tracking the state of a multi-protocol task becomes unreliable, which results in inconsistent execution outcomes. For instance, in a task where a data agent must first confirm data availability via A2A, a reporting agent might prematurely invoke a secure database tool via MCP before the data agent's A2A confirmation is finalized. This results in the generation of a report based on incomplete or unverified data.
\end{itemize}

\shortsectionBf{Coordination Issues.}
The absence of tools to validate the correctness of cross-protocol interactions creates design flaws, such as functional collapse and security breaches, that an adversary can exploit. First, \emph{circular delegation and deadlock} emerge when tasks pass indefinitely between agents. For example, Agent A delegates to Agent B (A2A), which in turn delegates back to Agent A (A2A), creating a cycle that halts progress. Second, \emph{privilege escalation} occurs when a malicious agent exploits delegation chains to gain unauthorized access to tools. For instance, Agent A, which lacks direct access to a sensitive MCP tool, delegates an innocent-sounding task to Agent B (A2A), which possesses the necessary MCP credentials, thereby using Agent B as a proxy for unauthorized access.

These challenges highlight the critical need for formal modeling frameworks that capture the essential properties of integrated agent-tool and agent-agent coordination. Such frameworks must be capable of reasoning about cross-protocol interactions, verifying end-to-end correctness properties, and detecting potential vulnerabilities before deployment in high-stakes environments.

\section{Modeling the Agentic AI Systems}
\label{sec:formal_models}
We abstract the formal model for the \emph{Host Agent} (\hostagent) to provide a unified view of an Agentic AI system. The \hostagent receives natural language requests from users, decomposes them into structured subtasks, assigns those subtasks to AI agents or other external entities, and aggregates the results. It also oversees execution, monitors progress, and manages exceptions or recovery procedures. We then complement the host agent model with the \emph{task lifecycle} model, which specifies the states and transitions a sub-task undergoes from its origin to completion. 
Our deliberate separation of the host agent and task lifecycle models serves two main purposes. First, it establishes clear abstraction boundaries to delineate responsibilities; the \hostagent handles orchestration and intent resolution, while the task model manages the execution details of individual sub-tasks. Second, it allows integration of new sub-task states to the task model without requiring fundamental changes to the \hostagent model.

We derive these models by synthesizing the operational structures, specifications, and reference implementations of the unified protocols (Anthropic's MCP and Google's A2A). This incorporates insights from modular orchestration and dynamic task planning practices for Agentic AI systems.


Recent orchestration frameworks such as LangGraph~\citep{langgraph}, n8n~\citep{n8n}, and OpenAI's AgentKit~\citep{OpenAI2025AgentKit} inspire the architectural principles of the \hostagent. However, these frameworks typically execute predefined, graph-based workflows. The proposed model extends this paradigm. The \hostagent creates and adjusts execution plans dynamically in response to user requests. This approach permits autonomous coordination and avoids the constraints of static flowcharts. Similarly, the task lifecycle model generalizes state management concepts from protocols such as A2A. This structure supports flexible task progressions. In this framework, runtime conditions and autonomous agent decisions (\eg failures and fallbacks) determine state transitions.

We leverage these models to abstract key operational structures of Agentic AI. This abstraction provides a foundation for property identification aimed at safe system design through the formalization of task delegation, inter-agent coordination, and execution oversight, detailed in Section~\ref{sec:formal_properties}.

\begin{figure}[t!]
\centering
    \includegraphics[width=.8\linewidth]{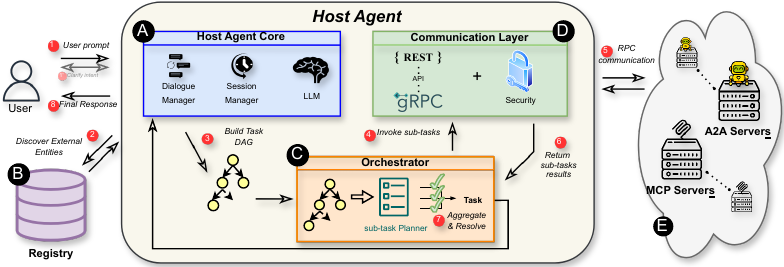}
    \caption{The Host Agent architecture.}
\label{fig:host_agent}
\vspace{-7pt}
\end{figure}

\subsection{Constructing the Host Agent Model}
\label{subsec:host_agent}
The Host Agent interprets users' queries, selects external entities, and manages task lifecycles (Figure~\ref{fig:host_agent}). It uses internal state, historical records, and external data to gather context for task decomposition and agent coordination.



\shortsectionBf{Definition of the Host Agent Model ($\mathcal{H}$).} We define the \hostagent model $\mathcal{H}$ as a tuple:
$\mathcal{H} = \left( \mathcal{A}, \mathcal{E}, \mathcal{T}, \mathcal{R}, \mathcal{C}, \mathcal{O}, \mathcal{C_{L}}, \mathcal{S}_{\mathcal{H}} \right)$
where the components are defined as follows:

\begin{itemize}
    \item $\mathcal{A}$: The set of all autonomous Agents (\eg A2A Servers).
    \item $\mathcal{E}$: The set of all External Entities (EEs), such that $\mathcal{A} \subseteq \mathcal{E}$, which includes functional tools (MCP).
    \item $\mathcal{T}$: The set of all possible user Tasks, where a task $T$ is a tuple $T = (Req_{U}, Resp_{H})$.
    \item $\mathcal{R}$: The Registry, which maintains a mapping from external entities to their capability profiles, $\mathcal{R}: \mathcal{E} \to \mathcal{P}(\mathcal{EE}_{info} \times \mathcal{M}_{API})$.
    \item $\mathcal{C}$: Host Agent Core (\hostagentcore), responsible for intent resolution $I_{U}$, $\mathcal{C}: Req_U \times \mathcal{S}_{SM} \to \mathcal{I}_{U}$.
    \item $\mathcal{O}$: Orchestrator, responsible for task decomposition, execution management, and aggregation.
    \item $\mathcal{C_{L}}$: Communication Layer (\communicationlayer), responsible for secure, reliable, and protocol-agnostic communication with External Entities in $\mathcal{E}$. 
    \item $\mathcal{S}_{\mathcal{H}}$: The state space of the host agent, encompassing the overall system state, including network activity and the status of all managed sub-tasks.
\end{itemize}

The Orchestrator's core function is task decomposition, mapping a resolved intent $I_U$ and available entities $\mathcal{E}$ to a Directed Acyclic Graph (DAG) of sub-tasks $\mathcal{D}$,
$\mathcal{O}_{\mathrm{decomp}}: \mathcal{I}_{U} \times \mathcal{R}(\mathcal{E}) \to \mathcal{D}$.
The execution management function, $\mathcal{O}_{\mathrm{exec}}$, maps the DAG through sub-task invocations (via $\mathcal{D}$) to the final results to be aggregated:
$\mathcal{O}_{\mathrm{exec}}: \mathcal{D} \to \mathcal{F}_{\mathrm{results}}$.

The \hostagent model provides the state space $\mathcal{S}_{\mathcal{H}}$ and critical functions ($\mathcal{C}, \mathcal{O}$) to define the liveness, safety, completeness, fairness, and reachability properties in Table~\ref{tab:host_formal_properties}.

\begin{table*}[!t]{
    \centering
    \scriptsize{
    \caption{Specification of Host Agent formal model properties ($\mathtt{HP_1}$--$\mathtt{HP_{16}}$).}
    \label{tab:host_formal_properties}
    \begin{tabular}{@{}p{0.05\textwidth}p{0.90\textwidth}@{}}
		\toprule \multicolumn{2}{c}{\textbf{Liveness}}\\
		\midrule 
		$\mathtt{HP_1}$ & \text{Every user prompt eventually receives a response from the \textsc{\scriptsize{HA}}.} \\
		& $AG(Req_{U} \rightarrow AF\: Resp_{H})$ \\
		\hdashline \addlinespace[2pt]
		$\mathtt{HP_2}$ & \text{The \textsc{\scriptsize{HAC}} eventually clarifies the intent of each user prompt.} \\
		& $AG(Req_{U} \rightarrow AF\: \mathcal{I}_{U})$ \\
		\hdashline \addlinespace[2pt]
		$\mathtt{HP_3}$ & \text{Once intent is clarified, the \textsc{\scriptsize{LLM}} eventually constructs a Task DAG using the discovered external entities.} \\
		& $AG(\mathcal{I}_{U} \rightarrow AF\: \mathsf{LLM.Task\_DAG}(\mathcal{I}_{U}, \{\mathcal{EE}_{info}\}))$ \\
		\hdashline \addlinespace[2pt]
		$\mathtt{HP_4}$ & \text{Whenever a Task DAG is successfully built, all sub-tasks it contains are eventually invoked.} \\
		& $AG(\mathsf{LLM.Task\_DAG}(\mathcal{I}_{U}, \{\mathcal{EE}_{info}\}) \rightarrow \forall i \in D:\mathsf{AF}\: (\mathsf{CL.invoke}(EE, prot, sub\_task_{i})))$ \\
		\hdashline \addlinespace[2pt]
		$\mathtt{HP_5}$ & \text{Every invoked sub-task eventually produces a result.} \\
		& $\mathsf{AG}\: (\mathsf{CL.invoke}(EE, prot, sub\_task) \rightarrow AF\: (\mathsf{CL.return\_result}(sub\_task)))$ \\
		\hdashline \addlinespace[2pt]
		$\mathtt{HP_6}$ & \text{Once all sub-task results are returned via the \textsc{\scriptsize{CL}}, the Orchestrator eventually aggregates and resolves them.} \\
		& $AG((\forall i \in D:\mathsf{HasResult}(sub\_task_i)) \rightarrow \mathsf{AF}(\mathsf{O.aggregate}(sub\_task_{1}, \dots, sub\_task_{n})))$ \\
		\midrule\midrule \multicolumn{2}{c}{\textbf{Safety}} \\
		\midrule 
		$\mathtt{HP_7}$ & \text{The Task DAG is constructed by the \textsc{\scriptsize{LLM}} only after entities have been discovered in the Registry.} \\
		& $AG \Big( \mathsf{LLM.Task\_DAG}(\mathcal{I}_{U}, \{\mathcal{EE}_{info}\}) \rightarrow \forall e \in \{\mathcal{EE}_{info}\}: \mathsf{is\_valid}(\mathsf{\mathcal{R}.discover}(e)) \Big)$ \\
		\hdashline \addlinespace[2pt]
		$\mathtt{HP_8}$ & \text{Sub-tasks are invoked only if the Task DAG $D$ has already been constructed and explicitly includes them.} \\
		& $AG(\mathsf{CL.invoke}(EE, prot, sub\_task)\rightarrow sub\_task \in D)$ \\
		\hdashline \addlinespace[2pt]
		$\mathtt{HP_9}$ & \text{Task invocation is strictly conditional on the EE having completed the system's predefined validation process.} \\
		& $AG \: (\mathsf{CL.invoke}(\text{EE}, \textit{prot}, \textit{payload}) \rightarrow VM(\text{EE}) )$ \\
		\hdashline \addlinespace[2pt]
		$\mathtt{HP_{10}}$ & \text{Every sub-task in the Task DAG $D$ is invoked only when it has no dependencies on other uncompleted sub-tasks.} \\
        & $AG \Big( \forall i \in D:\mathsf{CL.invoke}(EE, prot, sub\_task_{i}) \rightarrow \forall p \in \mathsf{parents}(sub\_task_{i}): \mathsf{Completed}(p) \Big)$ \\
		\hdashline \addlinespace[2pt]
		$\mathtt{HP_{11}}$ & \text{A response from the \textsc{\scriptsize{HA}} is returned to the user only after every corresponding sub-tasks have been invoked.} \\
		& $AG \Big( Resp_{H} \rightarrow (\forall i \in D: \mathsf{WasInvoked}(sub\_task_{i})) \Big)$ \\
		\midrule\midrule \multicolumn{2}{c}{\textbf{Completeness}} \\
		\midrule 
		$\mathtt{HP_{12}}$ & \text{Every user prompt leads either to intent clarification or to task planning.} \\
		& $AG(Req_{U} \rightarrow AF(\mathsf{LLM.Task\_DAG}(\mathcal{I}_{U}, \{\mathcal{EE}_{info}\}) \lor \text{Clarify\_Intent}))$ \\
		\midrule\midrule \multicolumn{2}{c}{\textbf{Fairness}} \\
		\midrule 
		$\mathtt{HP_{13}}$ & \text{All A2A agent RPC calls eventually receive responses (\ie do not remain pending indefinitely).} \\
		& $AG(\mathsf{A2A\_Call} \rightarrow AF(\mathsf{A2A\_Response}))$ \\
		\hdashline \addlinespace[2pt]
		$\mathtt{HP_{14}}$ & \text{JSON-RPC calls to MCP servers eventually succeed (\ie do not remain pending indefinitely).} \\
		& $AG(\mathsf{MCP\_Call} \rightarrow AF(\mathsf{MCP\_Response}))$ \\
		\midrule\midrule \multicolumn{2}{c}{\textbf{Reachability}} \\
		\midrule 
		$\mathtt{HP_{15}}$ & \text{It is always possible to reach a state in which the Host Agent replies to the user.} \\
		& $EF(Resp_{H})$ \\
		\hdashline \addlinespace[2pt]
		$\mathtt{HP_{16}}$ & \text{It is always possible to reach a state in which the \textsc{\scriptsize{LLM}} builds a Task DAG.} \\
		& $EF(\mathsf{LLM.Task\_DAG}(\mathcal{I}_{U}, \{\mathcal{EE}_{info}\}))$ \\
		\bottomrule
	\end{tabular}
}}
\end{table*}

\subsubsection{Functional Architecture and Execution Sequence.}
We detail the operation of \hostagent by referencing specific components and their interactions (\circled{A}-\circled{E}), as well as execution steps (\circledRed{1}-\circledRed{8}), as depicted in Figure~\ref{fig:host_agent}. This defines the role of each component and the data flow from the initial user prompt to the final response.

\shortsectionBf{Host Agent Core (\hostagentcore) - \circled{A}.}
The \hostagentcore is the first point of contact with users and processes their requests (\circledRed{1}). It serves as the central reasoning module and comprises three key sub-components:

\begin{itemize}
\item \textbf{Large Language Model (\llm)}: Central to the \hostagentcore's function, the \llm interprets user requests and generates coherent responses. The \llm and \dialoguemanager collaborate to resolve the underlying user intent ($\mathcal{I}_{U}$) upon receiving a request ($Req_U$).
\item \textbf{Session Manager (\sessionmanager)}: The \sessionmanager maintains dialogue context across multi-turn interactions and manages user-stored credentials for secure access to authenticated services. We integrate the \sessionmanager based on evidence that contextual management improves decision-making and reasoning in \llm-based systems~\citep{llmmultiagentsystemschallenges, tree-of-thoughts, memory-of-thought}.
\item \textbf{Dialogue Manager (\dialoguemanager)}: The \dialoguemanager handles conversation flow, working with the \llm to clarify ambiguous user intent and elicit necessary information before execution proceeds (\circledRed{1'}).
\end{itemize}

\noindent The \hostagent's context-aware reasoning resolves the user intent ($\mathcal{I}_{U}$), captured formally as $\text{HAC}(Req_U, \text{State}_{SM}) \to \mathcal{I}_{U}$, where $\text{State}_{SM}$ represents the current dialogue and credential state maintained by the \sessionmanager. Following intent resolution, the \hostagentcore initiates the subsequent task phases.

\shortsectionBf{Registry - \circled{B}.}
The Registry maintains a dynamic collection of \externalentity{s}--including autonomous AI agents (\ie A2A server), functional tools (\ie MCP servers), and other service endpoints (contained in \circled{E}). These entities are registered to expose capabilities leveraged during task execution. Each \externalentity requires a capability profile ($\mathcal{P}(\mathcal{EE}_{info} \times \mathcal{M}_{API})$), which details its skills ($\mathcal{EE}_{info}$) and associated API metadata ($\mathcal{M}_{API}$) necessary for invocation. The Registry supports entity registration, efficient querying, and deregistration to enable system extensibility. Post-resolution, the \hostagentcore leverages the Registry (\circledRed{2}) to discover suitable \externalentity{s}. Discovery is defined as $\mathsf{\mathcal{R}.discover}(EE) \to \mathcal{P}(\mathcal{EE}_{info} \times \mathcal{M}_{API})$, and entity registration as $\mathsf{\mathcal{R}.register}(\mathcal{P}(\mathcal{EE}_{info} \times \mathcal{M}_{API})) \to \textit{Success} \mid \textit{Failure}$.


\shortsectionBf{Orchestrator - \circled{C}.}
Using intent $\mathcal{I}_{U}$ and Registry entities, the \llm builds a DAG $\mathcal{D} = (\mathcal{V}, \mathcal{S})$ (\circledRed{3}) where nodes $\mathcal{V}$ are interrelated sub-tasks and edges $\mathcal{S}$ enforce precedence constraints or data dependencies. An edge $(v_i, v_j)$ indicates $v_j$ must await $v_i$'s completion. The Orchestrator schedules (\circledRed{4}), collects (\circledRed{6}), and aggregates (\circledRed{7}) results. This process is defined by $D := \mathsf{LLM.Task\_DAG}(\mathcal{I}_{U}, \{\mathcal{EE}_{info}\})$ and subsequent execution $\mathsf{O.execute}(D)$ where final output is aggregated $\forall i \in D: \mathsf{O.aggregate}(v_{1}, \dots, v_{n})$.

\shortsectionBf{Communication Layer (\communicationlayer) - \circled{D}.}
The \communicationlayer ensures secure, protocol-agnostic interaction with \externalentity{s} (\circled{E}), supporting standards such as REST and gRPC. It routes sub-tasks for invocation (\circledRed{5}) via $\mathsf{CL.invoke}(\text{EE}, \textit{prot}, \textit{payload}) \to \textit{resp}$. The $\text{payload}$ encapsulates API parameters, while $\textit{resp}$ returns results or errors to the Orchestrator. Finally, the \hostagentcore synthesizes the user output ($Resp_H$) via the \llm, delivering it through the \dialoguemanager (\circledRed{8}).





\label{subsec:task}
\begin{figure}[t!]
    \centering
    \includegraphics[width=.6\linewidth]{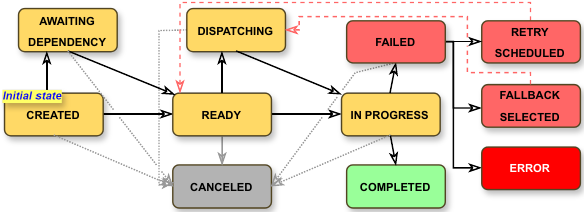}
    \caption{Sub-task state transitions including execution, delegation, recovery, and termination.}
    \vspace{-7pt}
    \label{fig:task}
\end{figure}

\subsection{Constructing the Task Lifecycle Model}
Building upon the \hostagent model, we introduce the task lifecycle model (Figure~\ref{fig:task}). It abstracts the sub-task lifecycle by detailing state transitions from origin to termination.

\shortsectionBf{Definition of the Task Lifecycle Model ($\mathcal{L}$).}
The task lifecycle model $\mathcal{L}$ defines the state space and transition function for a single sub-task $t$, where $t$ is a constituent node in the host agent's decomposition graph $D$.
Task lifecycle model $\mathcal{L} = \left( \mathcal{S}_{t}, s_{0}, \mathcal{E}_{t}, \delta \right)$:

\begin{itemize}
    \item $\mathcal{S}_{t}$: The set of discrete states a sub-task can occupy. This set is defined as: 
    
    $
    \mathcal{S}_{t} =
    \begin{medsize}
        \left\{
        \begin{matrix}
            \texttt{CREATED}, \texttt{AWAITING DEPENDENCY}, \texttt{READY},
            \texttt{DISPATCHING}, \texttt{IN PROGRESS}, \texttt{COMPLETED}, \\
            \texttt{FAILED}, \texttt{RETRY SCHEDULED},
            \texttt{FALLBACK SELECTED}, \texttt{CANCELED}, \texttt{ERROR}
        \end{matrix}
        \right\}
    \end{medsize}
    $
    \item $s_{0}$: The initial state, $s_{0} = \texttt{CREATED}$.
    \item $\mathcal{E}_{t}$: The set of external events and internal conditions (\eg dependency satisfaction, external failure signal, timeout) that trigger a state transition.
    \item $\delta$: The state transition function, $\delta: \mathcal{S}_{t} \times \mathcal{E}_{t} \to \mathcal{S}_{t}$. This function dictates the deterministic movement between states based on execution outcomes, protocol responses, and recovery policies.
\end{itemize}

The state space $\mathcal{S}_{t}$ and transition function $\delta$ are fundamental to verifying task execution integrity. They provide the basis for the safety invariants (\eg sequential transition checks) and liveness and fairness properties (\eg eventual termination/progression to $\texttt{READY}$ state), detailed in Section~\ref{sec:formal_properties}.

\shortsectionBf{Task and Sub-task Dynamics.}
In the lifecycle model, a task represents the high-level user request that serves as the core unit of work fulfilled by the \hostagent. A sub-task is a lower-level, constituent component into which the \hostagentcore decomposes the main task. Each sub-task has its own lifecycle, tracked within the parent task's context. This lifecycle with a defined sequence of states supports fine-grained control for asynchronous execution across distributed \externalentity{s}.

\shortsectionBf{Sub-task State Transitions.} The sub-task lifecycle begins in the \texttt{CREATED} state. Sub-task execution then involves three primary stages. For dependency management, if the sub-task requires other sub-tasks to complete (\ie incoming DAG edges exist), it enters the \texttt{AWAITING DEPENDENCY} state and remains paused until all prerequisites are satisfied. Once dependencies are satisfied, the sub-task transitions to the \texttt{READY} state. For execution initiation, a sub-task in the \texttt{READY} state becomes eligible for execution: the \hostagentcore delegates it to an external entity, transitioning to \texttt{DISPATCHING}, or the \hostagent handles it internally, proceeding directly to \texttt{IN PROGRESS}. For completion, successful execution (internal or external) moves the sub-task to the \texttt{COMPLETED} state, which makes its output available for aggregation and triggers dependent sub-tasks in the DAG. 

Execution is subject to failure and recovery mechanisms, where an execution failure causes the sub-task to transition immediately to the \texttt{FAILED} state. The system may then invoke recovery: the sub-task transitions to \texttt{RETRY SCHEDULED} if a retry mechanism is invoked. If retries are exhausted and a fallback entity is available, the state changes to \texttt{FALLBACK SELECTED}. A sub-task can also be set to \texttt{CANCELED} by the user or an \externalentity at various points in its lifecycle. If all recovery options fail, the sub-task reaches the \texttt{ERROR} state, which signifies terminal failure.

\begin{table*}[!th]
	\centering
	\scriptsize 
    \caption{Specification of Task Lifecycle formal model properties.}\label{tab:task_formal_properties}
    \begin{tabular}{@{}p{0.05\textwidth}p{0.90\textwidth}@{}}
		\toprule
		\multicolumn{2}{c}{\textbf{Liveness}} \\
		\midrule
		$\mathtt{TL_1}$ & \text{Every \texttt{CREATED} sub-task eventually terminates in \texttt{COMPLETED}, \texttt{ERROR}, or \texttt{CANCELED}.} \\
		& $AG (state = \texttt{CREATED} \rightarrow AF (state \in \{ \texttt{COMPLETED}, \texttt{ERROR}, \texttt{CANCELED} \} ))$ \\ \hdashline \addlinespace[2pt]
		
		$\mathtt{TL_2}$ & \text{A sub-task in \texttt{READY} that requires an external entity eventually transitions to \texttt{DISPATCHING}.} \\
		& $AG ((state = \texttt{READY} \land external\_entity\_needed) \rightarrow AF (state = \texttt{DISPATCHING}))$ \\ \hdashline \addlinespace[2pt]
		
		$\mathtt{TL_3}$ & \text{A sub-task in \texttt{FALLBACK SELECTED} eventually transitions to \texttt{DISPATCHING}, \texttt{CANCELED}, or \texttt{FAILED}.} \\
		& $AG (state = \texttt{FALLBACK SELECTED} \rightarrow AF (state \in \{ \texttt{DISPATCHING}, \texttt{CANCELED}, \texttt{FAILED} \} ))$ \\ \hdashline \addlinespace[2pt]
		
		$\mathtt{TL_4}$ & \text{A sub-task in \texttt{DISPATCHING} eventually reaches the \texttt{IN PROGRESS} state.} \\
		& $AG (state = \texttt{DISPATCHING} \rightarrow AF (state = \texttt{IN PROGRESS}))$ \\ \hdashline \addlinespace[2pt]

        $\mathtt{TL_{5}}$ & \text{A sub-task cannot remain indefinitely in \texttt{AWAITING DEPENDENCY}.} \\
		& $AG (state = \texttt{AWAITING DEPENDENCY} \rightarrow AF (state \neq \texttt{AWAITING DEPENDENCY}))$ \\ \hdashline \addlinespace[2pt]

        $\mathtt{TL_{6}}$ & \text{A sub-task in \texttt{AWAITING DEPENDENCY} with satisfied dependencies eventually transitions to \texttt{READY}.} \\
		& $AG (state = \texttt{AWAITING DEPENDENCY} \land dependencies\_satisfied \rightarrow AF (state = \texttt{READY}))$ \\
		
		\midrule\midrule
		\multicolumn{2}{c}{\textbf{Safety}} \\
		\midrule
		$\mathtt{TL_7}$ & \text{A sub-task enters \texttt{DISPATCHING} only from \texttt{READY}, \texttt{FALLBACK SELECTED}, or \texttt{RETRY SCHEDULED}.} \\
		& $AG ((state = \texttt{DISPATCHING}) \rightarrow previous\_state \in \{\texttt{READY}, \texttt{FALLBACK SELECTED}, \texttt{RETRY SCHEDULED}\})$ \\ \hdashline \addlinespace[2pt]
		
		$\mathtt{TL_8}$ & \text{A sub-task may only enter \texttt{COMPLETED} if it was previously \texttt{IN PROGRESS}.} \\
		& $AG ((state = \texttt{COMPLETED}) \rightarrow (previous\_state = \texttt{IN PROGRESS}))$ \\ \hdashline \addlinespace[2pt]
		
		$\mathtt{TL_9}$ & \text{Once a sub-task enters \texttt{ERROR}, it remains there permanently.} \\
		& $AG ( (state = \texttt{ERROR}) \rightarrow AX (state = \texttt{ERROR}) )$ \\ \hdashline \addlinespace[2pt]
		
		$\mathtt{TL_{10}}$ & \text{A sub-task may only enter \texttt{RETRY SCHEDULED} if its previous state was \texttt{FAILED}.} \\
		& $AG (state = \texttt{RETRY SCHEDULED} \rightarrow previous\_state = \texttt{FAILED})$ \\ \hdashline \addlinespace[2pt]
		
		$\mathtt{TL_{11}}$ & \text{A sub-task that enters \texttt{CANCELED} remains permanently in that state.} \\
		& $AG ( (state = \texttt{CANCELED}) \rightarrow AX (state = \texttt{CANCELED}) )$ \\
		
		\midrule\midrule
		\multicolumn{2}{c}{\textbf{Transition Constraints}} \\
		\midrule
		$\mathtt{TL_{12}}$ & \text{A \texttt{FAILED} sub-task with no fallbacks transitions to \texttt{RETRY SCHEDULED}, \texttt{ERROR} or is \texttt{CANCELED}.} \\
		& $AG ((state = \texttt{FAILED} \land \neg has\_fallbacks) \rightarrow AX (state \in \{\texttt{RETRY SCHEDULED}, \texttt{ERROR}, \texttt{CANCELED}\}))$ \\ \hdashline \addlinespace[2pt]
		
		$\mathtt{TL_{13}}$ & \text{A \texttt{FAILED} sub-task with fallbacks transitions to \texttt{RETRY SCHEDULED}, \texttt{FALLBACK SELECTED}, \texttt{ERROR} or is \texttt{CANCELED}.} \\
		& $AG ((state = \texttt{FAILED} \land has\_fallbacks) \rightarrow AX (state \in \{\texttt{RETRY}, \texttt{FALLBACK}, \texttt{ERROR}, \texttt{CANCELED}\}))$ \\ \hdashline \addlinespace[2pt]
		
		$\mathtt{TL_{14}}$ & \text{A sub-task in \texttt{RETRY SCHEDULED} transitions to \texttt{DISPATCHING} if the retry policy permits.} \\
		& $AG ((state = \texttt{RETRY SCHEDULED} \land retry\_policy\_permits) \rightarrow AX (state = \texttt{DISPATCHING}))$ \\
		\bottomrule
	\end{tabular}
\end{table*}

\section{Property Specification}
\label{sec:formal_properties}
The formalization of the \hostagent and \emph{Task Lifecycle} models provides the foundation for analyzing and verifying critical system properties. In this context, formal properties are verifiable statements about the system's execution paths, used to ensure the safety, security, and functionality of an Agentic AI system's behavior. We structure these properties into two distinct sets: those captured by the \hostagent model (overall orchestration, Table~\ref{tab:host_formal_properties}), and those governing the task lifecycle model (sub-task state transitions, Table~\ref{tab:task_formal_properties}). These properties do not operate in isolation; they form a tightly coupled system of interdependent guarantees necessary for correct and reliable function in a multi-AI agent environment. Both sets of properties are expressed using temporal logic, CTL and LTL, which allow for an unambiguous definition of desired system behavior and execution paths.

\subsection{Property Design Principles and Definitions}
Both models include a set of property categories, each containing multiple specific properties, to establish verifiable guarantees. The core categories used in both tables are liveness, safety, completeness, fairness, and reachability. These properties ensure the overall system makes progress, avoids undesirable states, achieves its goals, and handles reliable resource allocation.

\shortsectionBf{Liveness.} Liveness ensures that an Agentic AI system will eventually make progress toward a goal. This property guarantees that, despite asynchronous actions, inter-agent dependencies, or potential conflicts, the system avoids global deadlock or infinite starvation. This is critical here, where a lack of progress in one agent can propagate and result in a cascading stall of the entire system.

\begin{itemize}
    \item \textbf{Example ($\mathtt{HP_{1}}$):} Every user prompt must eventually receive a final response from the host agent. This is expressed in temporal logic as $AG(Req_{U} \to AF \ Resp_{H})$.
    \item \textbf{Example ($\mathtt{TL_{1}}$):} A sub-task that is $\texttt{CREATED}$ must eventually terminate in a defined end state ($\texttt{COMPLETED}$, $\texttt{ERROR}$, or $\texttt{CANCELED}$).
\end{itemize}

\shortsectionBf{Safety.} Safety ensures that agents never enter globally invalid or harmful states, even when operating asynchronously or under adversarial conditions. In an Agentic AI system, this includes verifying that an agent’s actions do not cause irreversible policy violations in the collective system state.

To formally reason about the safety of the system, we introduce the \emph{Validation Module} (\validationmodule). The \validationmodule is responsible for enforcing trust and validation constraints on external entities (\externalentity{s}). We define $VM(\text{EE})$ as the proposition that a given entity \externalentity has successfully passed all predefined validation processes, including schema validation and behavioral assessments. For the formal analysis, we assume the existence of a correct and functioning \validationmodule.

\begin{itemize}
    \item \textbf{Example ($\mathtt{HP{_9}}$):} This property ensures task invocation ($\mathsf{CL.invoke}$) is strictly conditional on the EE having completed system validation. Formally, $AG \: (\mathsf{CL.invoke}(\text{EE}, \textit{prot}, \textit{payload}) \rightarrow VM(\text{EE}) )$.
    \item \textbf{Example ($\mathtt{TL_{8}}$):} A sub-task may only enter the $\texttt{COMPLETED}$ state if its previous state was $\texttt{IN PROGRESS}$. This guards against incorrect state transitions, expressed as $AG ((state = \texttt{COMPLETED}) \rightarrow (previous\_state = \texttt{IN PROGRESS}))$.
    \item \textbf{Example ($\mathtt{HP_{10}}$):} A sub-task is invoked only when it has no uncompleted dependencies, preventing the consumption of intermediate data that is still being updated. This is expressed as $AG ( \forall i \in D:\mathsf{CL.invoke}(EE, prot, sub\_task_{i}) \rightarrow \forall p \in \mathsf{parents}(sub\_task_{i}): \mathsf{Completed}(p) )$.
\end{itemize}


\shortsectionBf{Completeness.} It captures the guarantee that if a valid solution or coordinated plan exists, the multi-AI agent system will find it. This property is relevant for distributed planning, where incomplete reasoning could leave solvable tasks unfulfilled due to suboptimal agent coordination.
\begin{itemize}
    \item \textbf{Example ($\mathtt{HP_{12}}$):} Every user prompt must lead either to intent clarification or to task planning, ensuring the system processes every request and initiates a path toward a solution. This behavior is captured as $AG(Req_{U} \rightarrow AF(\mathsf{LLM.Task\_DAG}(\mathcal{I}_{U}, \{\mathcal{EE}_{info}\}) \lor \text{Clarify\_Intent}))$.
\end{itemize}

\shortsectionBf{Fairness.} Fairness requires that all external operations delegated to A2A agents or MCP servers must eventually terminate, either by returning a result or triggering a failure condition (\eg timeout). This assumption prevents the Host Agent from blocking indefinitely on unresponsive external entities, ensuring that the main control loop remains live.
\begin{itemize}
    \item \textbf{Example ($\mathtt{HP_{13}}$):} Any RPC call initiated to an external A2A agent must eventually receive a response. This guarantees that network or agent-level delays do not cause system-wide deadlock and is expressed as $AG(\mathsf{A2A\_Call} \rightarrow AF(\mathsf{A2A\_Response}))$.
\end{itemize}

\shortsectionBf{Reachability.} Reachability requires that desired joint states or configurations are achievable through the agents’ available actions and communication patterns. In multi-AI agent coordination, this involves verifying that the combination of capabilities and allowed transitions can lead from the current distributed state to the intended goal without deadlocks or inescapable loops.
\begin{itemize}
    \item \textbf{Example ($\mathtt{HP_{15}}$):} It is always possible to reach a state where the host agent replies to the user. This property is expressed as $EF(Resp_{H})$.
\end{itemize}

\section{Case Study: Verification of Architectural Controls against Adversarial Behavior}
\label{sec:adversary}

The \hostagent and task lifecycle models provide a framework to detect, constrain, and mitigate adversarial behaviors in Agentic AI. Here, adversarial behavior is defined as any external deviation that compromises correctness, violates system invariants, or exploits control flow inconsistencies. These models support a layered security architecture, where security constraints are encoded as temporal logic invariants at distinct architectural layers. This structure enables verification of system-level defenses, including secure communication, trust anchoring, intent integrity, and failure containment.

\shortsectionBf{Control Point 1: Host Agent Core (Intent Integrity).} The \hostagentcore serves as the primary human-AI interface and the initial security boundary. This layer is vulnerable to prompt injection~\citep{promptinjectionattackllmintegrated, indirect_PI, llm_to_llm_PI} and jailbreak attacks~\citep{jailbroken} that attempt to subvert user intent. Our model establishes an extensible security boundary by leveraging the explicit ``Clarify Intent'' phase (\circledRed{1'}) as a guard. Correctness and completeness are enforced via property $\mathtt{HP_{12}}$, which requires every user request to induce a traceable execution path toward clarification or planning, ensuring no request is silently discarded. Additionally, progress and liveness are guaranteed by property $\mathtt{HP_{2}}$, asserting that the core eventually clarifies the intent of each request ($AG(Req_{U} \rightarrow AF\: \mathcal{I}_{U})$), thereby preventing denial-of-service from indefinite ambiguity.

\shortsectionBf{Control Point 2: Registry (Trust Anchoring).} The Registry functions as the trust anchor for all \externalentity interactions, mitigating supply-chain risks from malicious integrations~\citep{mcp_threats, mcp_urgently_privilege}. Trust soundness is enforced by safety property $\mathtt{HP_{9}}$, which constrains task invocation such that any $\mathsf{CL.invoke}$ call is permitted only if the target \externalentity has successfully completed system validation ($VM(\text{EE})$). This formal verification ensures runtime adherence to established trust requirements, preventing privilege escalation by unvetted entities.

\shortsectionBf{Control Point 3: Orchestrator (Delegation Monitoring).} The Orchestrator acts as a runtime enforcement mechanism monitoring task delegation through a dependency DAG to defend against coordination-based threats. Execution integrity and ordering are ensured by property $\mathtt{HP_{10}}$, requiring that task invocation occurs only after all unresolved dependencies reach a terminal success state. Furthermore, the Orchestrator enforces causal isolation and failure containment via dependency management, confining adversarial effects to the minimal execution subgraph (e.g., a sub-task does not proceed if any dependency is \texttt{FAILED}), thereby containing fault propagation.

\shortsectionBf{Control Point 4: Communication Layer and Zero-Trust.} The \communicationlayer provides a protocol-agnostic security substrate enforcing a zero-trust model. Protocol authenticity and integrity are preconditions for every $\mathsf{CL.invoke}$ operation, requiring verifiable identity and message confidentiality before processing payloads. Reliability and fairness are captured by properties $\mathtt{HP_{13}}$ and $\mathtt{HP_{14}}$, which require that all A2A and MCP RPC invocations eventually receive a response, verifying the continuous availability of the communication infrastructure under adversarial conditions.

\section{Related Work}

\shortsectionBf{Multi-Agent System (MAS) Architectures and Formal Models.}
Traditional MAS research focused on authorization and secure communication~\citep{security_in_mas2015}, but modern Agentic AI requires new coordination frameworks. While recent surveys provide taxonomies for these systems~\citep{llm-mas-survey, modeling-detecting-Intention, llm-agents-survey}, they lack operational models for property verification. We advance this by formalizing Host Agent and Task lifecycle models to prove end-to-end correctness. Unlike empirical studies that only demonstrate failure, our property specification approach--grounded in rigorous logic--offers a mathematically verifiable solution for reliability.

\shortsectionBf{The Challenge of Integrated Protocol Security.}
Agentic AI security is a major domain motivating our framework. Documented challenges include coordination attacks, leakage, and privilege escalation via delegation~\citep{dewitt2025openchallengesmultiagentsecurity}. These highlight cascade failures, where single-agent compromises propagate system-wide. Industry assessments such as OWASP's draft ``Top 10 for Agentic AI'' cite ``Unreliable Delegation \& Coordination'' as a critical risk, directly relating to the heterogeneous protocol integration we address~\citep{owaspagenticai}.

\shortsectionBf{Protocol-Specific Vulnerabilities.}
Rapidly emerging protocol vulnerabilities underscore the need for formal verification. MCP research identifies threats such as installer spoofing, sandbox escapes, and privilege gaps~\citep{mcp_threats, mcp_urgently_privilege}. Studies show \llm{s} can be coerced via MCP tools, enabling deception attacks and necessitating security benchmarks~\citep{radosevich2025mcpsafetyauditllms, mcp_sec_bench}. Similarly, A2A analysis reveals identity and task exchange vulnerabilities~\citep{buildingsecurea2a}. Key risks include message integrity and authentication, requiring validation to mitigate manipulation~\citep{neelou2025a2as}. Critical threats (e.g., prompt infection~\citep{llm_to_llm_PI}) allow malicious instructions to propagate between agents, persisting across tasks and potentially triggering system-wide cascades. Empirical studies confirm malicious agents can exploit delegation to subvert goals, proving traditional testing insufficient for cross-protocol interactions~\citep{motwani2024secret, combination_misuse_adversaries, redteamingllmmas}. This confirms the urgent need for a unified framework verifying integrated A2A and MCP interactions.

\section{Conclusion}
The fragmentation of protocols in Agentic AI prevents rigorous analysis of system safety, security, and functionality. We address this semantic gap by introducing a framework modeling the Host Agent and task lifecycle. Grounded in these models, we define 30 temporal logic properties, including liveness, safety, completeness, and fairness. These properties enable formal verification, edge-case detection, and deadlock prevention, which offer a domain-agnostic approach for designing verifiably safe systems. Future work will operationalize this methodology by automatically deriving formal models from code to check for property violations.

\subsubsection*{Acknowledgments}
This material is based upon work supported by the National Science Foundation under grant no. 2229876 and is supported in part by funds provided by the National Science Foundation (NSF), by the Department of Homeland Security, and by IBM. Any opinions, findings, and conclusions or recommendations expressed in this material are those of the author(s) and do not necessarily reflect the views of the NSF or its federal agency and industry partners.

\bibliography{references}
\bibliographystyle{iclr2026_conference}

\end{document}